%% file: main.tex
\documentclass[10pt,twocolumn,letterpaper]{article}

 \usepackage{cvpr}              

\input{preamble}

\definecolor{cvprblue}{rgb}{0.21,0.49,0.74}
\usepackage[pagebackref,breaklinks,colorlinks]{hyperref}

\newcommand{\cmark}{{\color{green}\ding{51}}}%
\newcommand{\xmark}{{\color{red}\ding{55}}}

\def\paperID{316} 
\def\confName{3DV\xspace}
\def\confYear{2024\xspace}

\title{Relative Pose for Nonrigid Multi-Perspective Cameras: The Static Case}

\author{Min Li\thanks{ two authors contribute equally.}\\
ShanghaiTech University\\
Shanghai, China\\
{\tt\small limin1@shanghaitech.edu.cn}
\and
Jiaqi Yang\footnotemark[1]\\
ShanghaiTech University\\
Shanghai, China\\
{\tt\small yangjq@shanghaitech.edu.cn}
\and
Laurent Kneip\\
ShanghaiTech University\\
Shanghai, China\\
{\tt\small lkneip@shanghaitech.edu.cn}
}

\begin{document}
\maketitle
\input{sec/0_abstract}    
\input{sec/1_intro}
\input{sec/2_relatedwork}
\input{sec/3_theory}
\input{sec/4_syntheticExp}
\input{sec/5_realExp}
\input{sec/6_discussion}
{
    \small
    \bibliographystyle{ieeenat_fullname}
    \bibliography{main}
}

\end{document}

%% file: preamble.tex
\usepackage[dvipsnames]{xcolor}

\usepackage{times}
\usepackage{epsfig}
\usepackage{graphicx}
\usepackage{epstopdf}
\usepackage{amsmath}
\usepackage{amssymb}
\usepackage{pifont}
\usepackage{multirow}

%% file: sec/0_abstract.tex
\begin{abstract}
   
   Multi-perspective cameras with potentially non-overlapping fields of view have become an important exteroceptive sensing modality in a number of applications such as intelligent vehicles, drones, and mixed reality headsets. In this work, we challenge one of the basic assumptions made in these scenarios, which is that the multi-camera rig is rigid. More specifically, we are considering the problem of estimating the relative pose between a static non-rigid rig in different spatial orientations while taking into account the effect of gravity onto the system. The deformable physical connections between each camera and the body center are approximated by a simple cantilever model, and inserted into the generalized epipolar constraint. Our results lead us to the important insight that the latent parameters of the deformation model, meaning the gravity vector in both views, become observable. We present a concise analysis of the observability of all variables based on noise, outliers, and rig rigidity for two different algorithms. The first one is a vision-only alternative, while the second one makes use of additional gravity measurements. To conclude, we demonstrate the ability to sense gravity in a real-world example, and discuss practical implications.

\end{abstract}

%% file: sec/1_intro.tex
\vspace{-0.7 cm}
\section{Introduction}
\vspace{-0.2 cm}
\label{sec:intro}
Multi-perspective cameras (MPCs) are considered an interesting exteroceptive sensor alternative for a variety of intelligent mobile systems such as drones~\cite{skydio}, intelligence augmentation devices~\cite{hololens}, autonomous ground vehicles~\cite{sevensense}, and---most prominently---self-driving cars~\cite{tesla,heng19}. Though limited overlap between the different cameras often limits the availability of intra-camera correspondences, it also leads to the advantage of omni-directional sensing and thus a better ability to---for example---recover accurate relative camera displacement information. Relative pose estimation with non-overlapping MPCs is the topic of this paper.
\begin{figure}[t]
\centering
  \includegraphics[width=0.9\columnwidth]{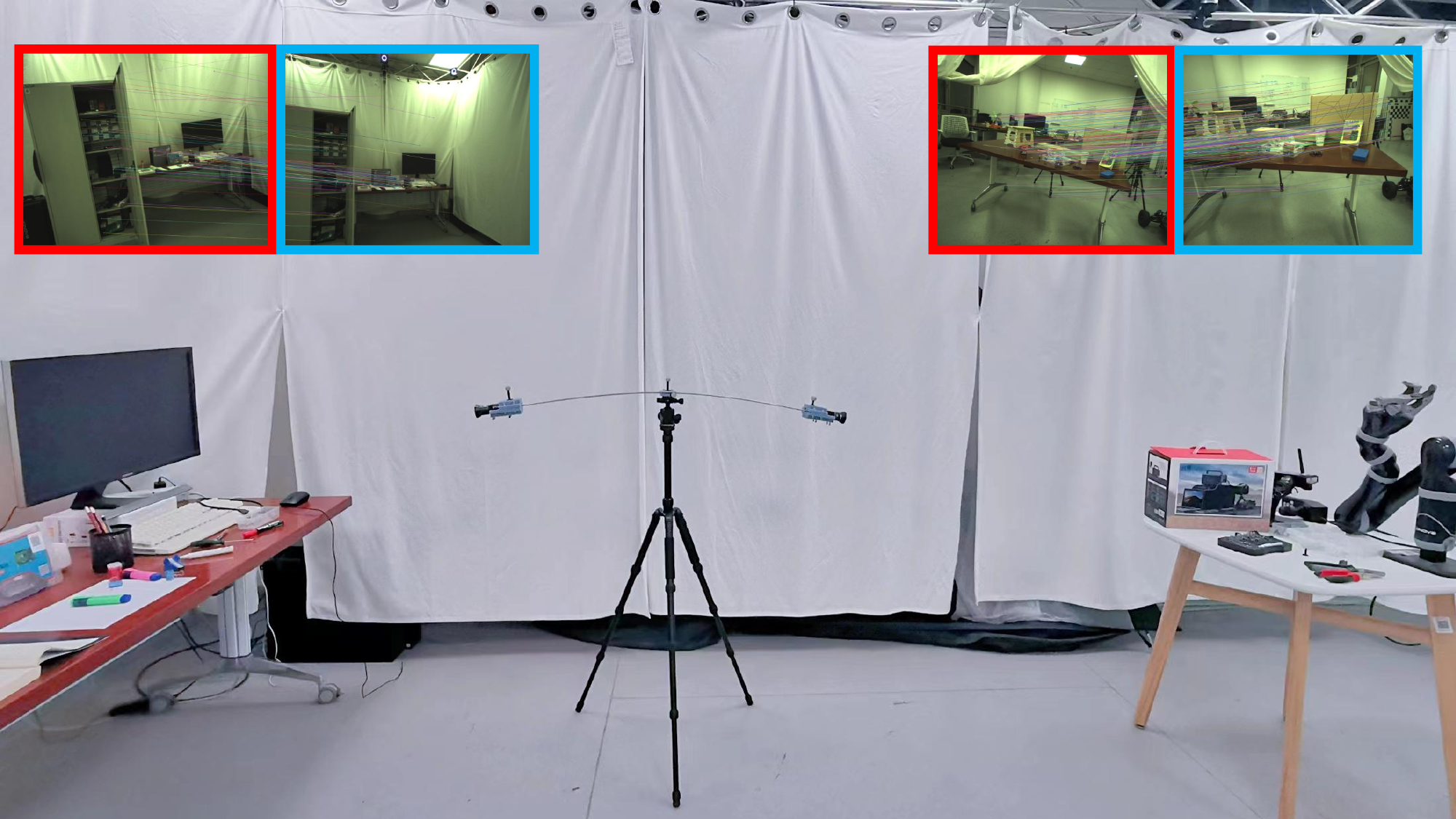}
\vspace{-0.2 cm}
  \caption{Illustrations of Nonrigid multi-perspective cameras.}
\vspace{-0.5 cm}
  \label{fig:real_experiment_setup}
  \vspace{-0.18 cm}
\end{figure}

\begin{table*}[t]
 \caption{Overview of different camera models presented over time. The presently proposed NR-MPC includes a physics-based deformation model enabling the observation of forces acting onto the body.\label{table:overview}}
\vspace{-0.3 cm}
 \centering
 \begin{tabular}{| l | c c c c |}
 \hline
 & omni-directional & metric & non-rigid & physics-based \\
 \hline
 Perspective camera & \xmark & \xmark & \xmark & \xmark \\
 Omni-directional camera & \cmark & \xmark & \xmark & \xmark \\
 Multi-Perspective Camera (MPC)~\cite{pless03} & \cmark & \cmark & \xmark & \xmark \\
 Articulated Multi-Perspective Camera (AMPC)~\cite{peng19} & \cmark & \cmark & \cmark & \xmark \\
 Non-Rigid Multi-Perspective Camera (NR-MPC) & \cmark & \cmark & \cmark & \cmark \\
 \hline
 \end{tabular}
 \vspace{-0.5 cm}
 \end{table*}

The exterior orientation estimation problem with non-overlapping MPCs is commonly solved using Pless' paradigm of \textit{using many cameras as one}~\cite{pless03}. By using known intrinsics and known camera-to-body extrinsic transformations, all image measurements can be converted into spatial rays expressed in one common body frame. Also known as the \textit{generalized camera model}, this form permits the expression of image correspondences via pairs of Pl\"ucker line vectors~\cite{plucker65}, which then constrain the \textit{generalized essential matrix}.

While the resulting solutions to the generalized relative pose problem have led to substantial advances, the present paper touches on one of the basic assumptions made in practically all of the related literature: \textit{rigid extrinsic transformations}. 
Let us start by considering a few practically relevant scenarios in which rigidity is not given.
For example, the structure of light-weight MAVs\cite{croon_remes_2016} and fixed-wing UAVs\cite{hinzmann18} undergoes substantial deformations as the system is exposed to forces. Within the robotics community, there is currently a growing interest in inherently safe, soft robotic manipulators\cite{9981335,sut2023}. Finally, heavy point masses on sensor rigs (e.g. the sensors themselves) may ultimately bend the support structure(cf. Figure \ref{fig:real_experiment_setup}).

The present paper studies the basic case in which the sensor setup in each pose is static and simply exposed to gravity. The analysis is furthermore conducted based on a demonstrative physical model in which all cameras are connected to the rig center via a bar-like structure that can be modeled by the \textit{cantilever model}\cite{Hibbeler2017,Shan2015}. While not necessarily corresponding to a commonly used physical system, the model serves well to analyze basic properties of observability within the context of the generalized relative pose estimation problem. We make the following contributions:
%
\begin{itemize}
  \item We express the extrinsic transformations between camera and rig body frames as a function of a physical deformation model that depends on the direction of gravity. In order to form the generalized epipolar constraint, these parametric extrinsics are then used to formulate the Pl\"ucker line coordinates. Note that the latter are no longer a pure function of measurements, but depend additionally on the unknown direction of gravity.
  \item Based on this incidence constraint, we propose two new algorithms to jointly find the generalized relative pose as well as the direction of gravity in both views. Taking redundancies into account, the new problem has eight degrees of freedom. The first algorithm is purely vision-based and considers the new model in the final optimization stage, while the second algorithm relies on an additional IMU prior to respect the deformable extrinsics already at the robust model fitting stage. The inclusion of the deformation model is challenging as it generally augments the dimensionality and non-linearity of the underlying objectives.
  \item We prove the perhaps unintuitive fact that even in the absence of direct measurements of either gravity or intra-camera correspondences, it is possible to constrain the deformation model and observe the orientation of the rig with respect to gravity as a latent variable besides the regular euclidean relative pose. The validation is carried out primarily on simulation experiments that confirm the observability of the latent parameters as well as investigate how material properties (e.g. stiffness) will influence signal-to-noise ratio. A successful application to a real-world case is provided as well.
\vspace{-0.2em}
\end{itemize}

While the present paper only considers the static case based on a slightly artificial deformation model, we believe that these results make an important statement. To the best of our knowledge, we believe that we show for the first time that the inclusion of a physical self-deformation model into geometric computer vision not only increases accuracy, but potentially reveals inertial states that would otherwise only be measurable by the addition of prior knowledge about the environment (e.g. knowledge about vertical lines) or a direct Inertial Measurement Unit. We define a new type of camera, denoted \textit{Non-Rigid Multi-Perspective Camera (NR-MPC)}, which takes into account the physical properties of the support structure to model deformation. A comparison between different camera models and their properties is listed in Table\ref{table:overview}. A discussion on the spectrum of practical applicability of the current method as well as potential down-stream implications of NR-MPCs is included at the end of the paper.

%% file: sec/2_relatedWork.tex
\vspace{-0.3 cm}
\section{Related Work}
\label{sec:relatedwork}
\vspace{-0.2 cm}
Two-view geometry is one of the most fundamental problems in geometric vision, and solutions traditionally find the essential matrix~\cite{higgins81}. Hartley~\cite{hartley97} adds normalization, thereby stabilizing the numerical accuracy of the algorithm. Nister~\cite{nister04}, Stewenius et al.~\cite{stewenius06} and Kukelova et al.~\cite{kukelova08} later on introduce the first minimal solutions to the problem. More recently, Kneip and Lynen~\cite{kneip13} introduce a rotation-only solver, and Briales et al.~\cite{briales18} introduce the first certifiably globally optimal solver to the non-minimal case. Yang et al.~\cite{jiaoling14} propose a correspondence-less solver able to find the global optimum despite the presence of outliers.

The present paper looks at generalizations of the problem finding the relative pose between two sets of extrinsically calibrated cameras. Pless~\cite{pless03} introduces the generalized essential matrix, which enables the first linear solutions. Stewenius and Nister~\cite{stewenius05} lateron introduce the first minimal solver to the problem, which is also used in this work. Kim et al.~\cite{kim07} present an alternative solution based on second-order cone-programming. Further advancements that discover degenerate cases of the linear 17-point algorithm and extend its applicability are presented by Li et al.~\cite{li08} and Kim and Kanade~\cite{kim10}. Zhao et al.~\cite{zhao20} propose the first  certifiably globally optimal solution to the non-minimal case.

The community has since ever put particular emphasis on the problem of non-overlapping multi-camera motion estimation, a problem in which no intra-camera correspondences are given and inter-camera correspondences are measured by the same camera in both views. Clipp et al.~\cite{clipp08} propose the first solution which uses five points in one view, and an additional point in the second view to resolve scale. Kim et al.~\cite{kim10b} propose multiple linear and L$\infty$ solutions. Kazik et al.~\cite{kazik12} propose a complete visual odometry system solving robustly for scale over multiple views. Kneip and Li~\cite{kneip14} again propose a solver that directly optimizes the relative rotation. Wang et al.~\cite{wang17,wang20} further investigate the problem of scale recovery, and Heng et al.~\cite{heng19} propose a complete stack of solutions for autonomous driving with surround-view camera systems.

Further categories of relative pose solvers make use of special motion parametrizations~\cite{scaramuzza09,scaramuzza09b,lee13,huang19,gao20}, or solvers that make use of a known directional correspondence~\cite{fraundorfer10,sweeney14}. A directional correspondence may for example come from a vanishing point correspondence, or otherwise direct gravity measurements. The solver of Sweeney et al.~\cite{sweeney14} is used in the present work for the case of known gravity measurements.

Note that a relatively complete collection of relative pose solvers including solvers for the multi-camera case are provided through the open-source library OpenGV~\cite{opengv}. Note furthermore that---though perhaps less relevant---there is a further generalization of the problem, which involves solving for a similarity transformation rather than just a Euclidean relative pose~\cite{sweeney15,kneip16}.

A highly related work to ours is given by Peng et al.~\cite{peng19}, who propose a relative pose solver for the case of an articulated multi-perspective camera. Perhaps the most related work to ours is presented by Hinzmann et al.~\cite{hinzmann18}, as they propose a stereo vision algorithm that takes into account a mechanical deformation model for the support structure of the cameras. Note however that their algorithm is specialized to the case of stereo vision, and does not address the more challenging case of non-overlapping relative pose estimation, nor does it touch on the possibility of recovering inertial readings from the deformation. To the best of our knowledge, our work is the first to consider this case.

%% file: sec/3_theory.tex
\vspace{-0.3 cm}
\section{Theory}
\label{sec:theory}
\vspace{-0.1 cm}
\subsection{Review of Relative Multi-Perspective Camera Geometry}
\vspace{-0.2 cm}
A Multi-Perspective Camera (MPC) is an ideally rigid construct of $M$ cameras adhering to a central projection model (i.e. each set of projection rays from an individual camera is intersecting in a single point). Let the pose of the $M$ cameras be defined in a fixed body frame $\mathcal{B}$ (the rig support frame). Let $\mathbf{v}_i,i=1,\ldots,M$ be the positions of the $M$ cameras expressed in $\mathcal{B}$, and let $\mathbf{V}_i,i=1,\ldots,M$ be their orientations which can be used to rotate points from the respective camera frame into the body frame $\mathcal{B}$. We furthermore assume that the cameras are intrinsically calibrated, and that we know the functions $\mathbf{u} = \pi_i(\mathbf{x})$ that transform 3D points $\mathbf{x}$ expressed in the $i$-th camera frame into image points $\mathbf{u}$. We furthermore know the functions $\mathbf{f} = \pi_i^{-1}(\mathbf{u})$ that transform image points back into unit-norm 3D spatial direction vectors expressed inside the camera frames. Note that depth gets lost along the projection into the image plane and $\mathbf{f} = \pi_i^{-1}(\pi(\mathbf{x})) = \frac{\mathbf{x}}{\|\mathbf{x}\|}$, hence $\pi_i^{-1}(\cdot)$ is not the exact inverse function of $\pi_i(\cdot)$. Note furthermore that we make no assumptions on $\pi_i(\cdot)$ other than that it models a central projection between 3D and 2D. Our theory therefore remains applicable to cameras with wide-angle fields of view. The assumption of a central projection is required to normalize measurements and express them by a direction vector $\mathbf{f}$ seen from a fixed origin.

\begin{figure}[t]
  \centering
  \includegraphics[width=\columnwidth]{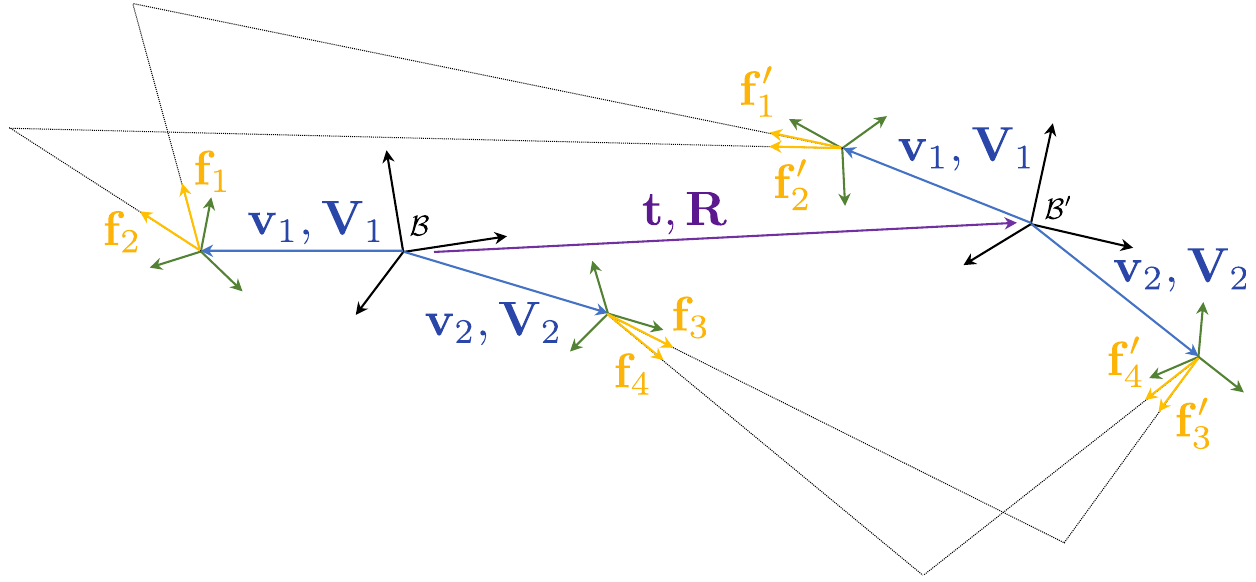}
\vspace{-0.7 cm}
  \caption{Geometry of the traditional, rigid MPC problem. Points are observed in two different body frame poses $\mathcal{B}$ and $\mathcal{B}'$ by cameras with extrinsics $(\mathbf{v}_j,\mathbf{V}_j)$. Example camera frames are indicated in green, and direction vectors $\mathbf{f}$ are all expressed inside these frames.}
  \label{fig:rigidSetup}
\vspace{-0.6 cm}
\end{figure}

We now assume that the MPC is measuring images in two different poses referred to as $\mathcal{B}$ and $\mathcal{B}'$. The relative displacement is given by $\mathbf{R}$ and $\mathbf{t}$ defined such that it is a Euclidean transformation of points from $\mathcal{B}'$ to $\mathcal{B}$. We furthermore assume that we have $N$ pairs of 3D point measurements---called \textit{correspondences}---where the first element always refers to a measurement by a camera in $\mathcal{B}$, and the second element refers to a measurement by the same camera in $\mathcal{B'}$. On one hand, we assume that the fields of view of the same camera in $\mathcal{B}$ and $\mathcal{B}'$ have significant overlap, hence corresponding inter-camera measurements between $\mathcal{B}$ and $\mathcal{B}'$ are made by the same camera. On the other hand, owing to the most general assumption of non-overlapping fields of view between the different cameras in one MPC pose, there are no intra-camera correspondences. Each correspondence pair can therefore be expressed by the tuple $\{ \mathbf{f}_i, \mathbf{f}'_i, j_i \}$, where $\mathbf{f}_i$ is the camera-referenced direction vector in $\mathcal{B}$, $\mathbf{f}'_i$ the corresponding direction vector in $\mathcal{B}'$, and $j_i\in \{1,\ldots,M\}$ the index of the camera that measured the $i$-th correspondence. The geometry of multi-perspective cameras is visualized in Figure~\ref{fig:rigidSetup}.

As introduced by Pless~\cite{pless03}, the incidence relation between rays in frame $\mathcal{B}$ and $\mathcal{B}'$ can now be stated as
\vspace{-0.2 cm}
\begin{equation}
  \mathbf{l}^T_i \mathcal{G} \mathbf{l}'_i = 0 \text{, where }
  \label{eq:generalizedepipolarconstraint}
\end{equation}
\vspace{-0.6 cm}
\begin{equation}
    \mathbf{l}_i  =  \left[\begin{matrix} \left(\mathbf{V}_{j_i}\mathbf{f}_i\right)^T  \left(\mathbf{v}_{j_i} \times \left(\mathbf{V}_{j_i}\mathbf{f}_i\right) \right)^T \end{matrix} \right]^T  
\end{equation}
\vspace{-0.4 cm}
\begin{equation}
    \mathbf{l}'_i  =  \left[\begin{matrix} \left(\mathbf{V}_{j_i}\mathbf{f}'_i\right)^T  \left(\mathbf{v}_{j_i} \times \left(\mathbf{V}_{j_i}\mathbf{f}'_i\right) \right)^T \end{matrix} \right]^T 
\end{equation}
are the Pl\"ucker line coordinates~\cite{plucker65} helping to express each measurement in $\mathcal{B}$ and $\mathcal{B}'$ as a spatial ray, and
\vspace{-0.2 cm}
\begin{equation}
    \mathcal{G} = \left[ \begin{matrix} \lfloor \mathbf{t} \rfloor_{\times} \mathbf{R} & \mathbf{R} \\ \mathbf{R} & \mathbf{0} \end{matrix} \right] 
    \label{eq:generalizedessentialmatrix}
\end{equation}
is what we call the \textit{generalized essential matrix}.

\vspace{-0.1 cm}
Note that---while the present introduction assumes a multi-perspective camera and thus measurements made by similarly placed cameras in $\mathcal{B}$ and $\mathcal{B}'$---the introduced theory remains mostly valid for generalized cameras where every 3D point observation in either $\mathcal{B}$ or $\mathcal{B}'$ is considered an independent measurement captured by a camera with arbitrary position and orientation inside the body frame. Also note that, while the figures show two-camera systems and the experiments are conducted on two-camera and four-camera systems, generalized epipolar geometry works with any number of cameras.

\subsection{Expressing the Static Case Deformations with the Cantilever Model}
\vspace{-0.1 cm}
As mentioned upfront, the present paper considers the static non-rigid case in which the camera setup is simply exposed to gravity. The following assumptions are made:
\begin{itemize}
    \item The Non-Rigid Multi-Perspective Camera (NR-MPC) is static and---without lack of generality---suspended in the body centers $\mathcal{B}$ and $\mathcal{B}'$.
    \item We model the connection of each camera to the body center by an elastic bar. We furthermore assume that the only masses in the system are point masses that are located in the camera centers or in the suspended body center. While not entirely general, this model serves well to explore basic properties of observability in the static case. Furthermore, there are important practical cases that come close to this model, for example the case of camera rigs where the individual sensors are connected to a common base via light-weight bars, i.e., light-weight MAVs\cite{croon_remes_2016} and fixed-wing UAVs\cite{hinzmann18}.
\end{itemize}

\begin{figure}[t]
  \centering
  \vspace{-0.3 cm}
  \includegraphics[width=\columnwidth]{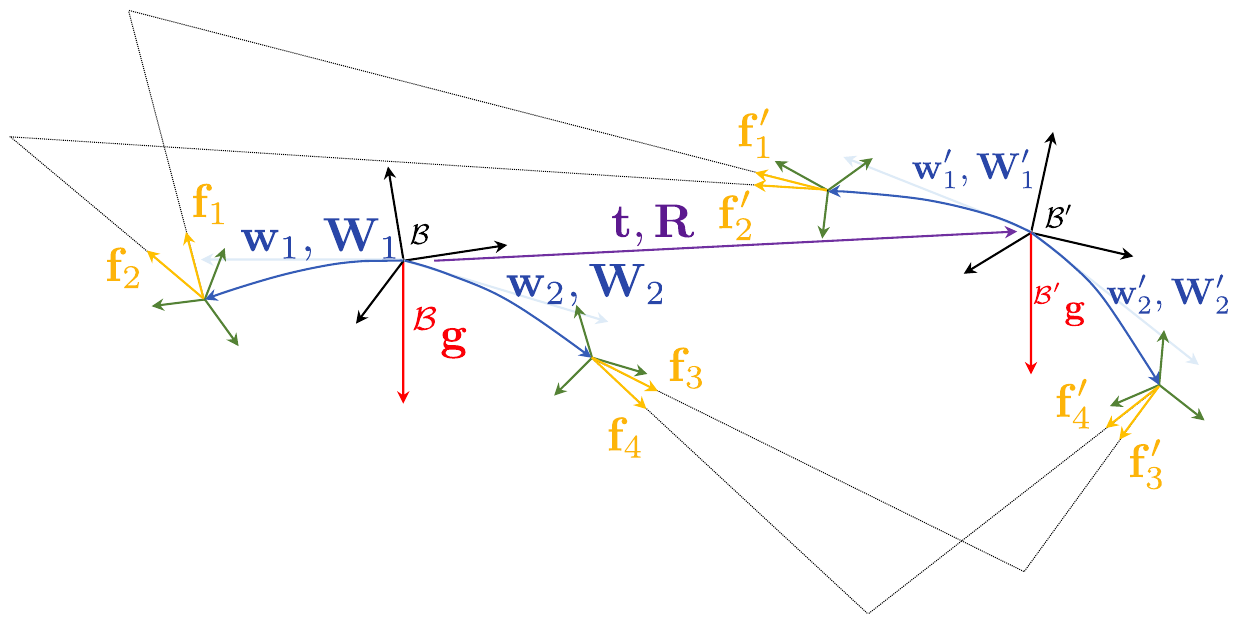}
  \vspace{-0.8 cm}
  \caption{Case of an NR-MPC where the bars connecting to the body center are deforming under the influence of gravity.}
 \vspace{-0.6 cm}
  \label{fig:nonrigidSetup}
\end{figure}

The non-rigid setup is explained in Figure~\ref{fig:nonrigidSetup}, the difference being that now we have variable extrinsic transformations $(\mathbf{w}_{j_i},\mathbf{W}_{j_i})$ and $(\mathbf{w}'_{j_i},\mathbf{W}'_{j_i})$ for the $j_i$-th camera that perceived the $i$-th correspondence. In the following, we will derive a physical model describing these extrinsics based on the orientation of the rig with respect to gravity.

\vspace{-0.1 cm}
\begin{figure}[b]
  \centering
\vspace{-0.3 cm}
  \includegraphics[width=0.4\columnwidth]{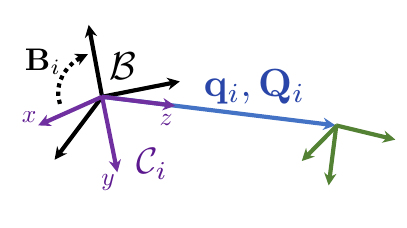}
\vspace{-0.4 cm}
  \caption{Location and orientation of the intermediate frame $\mathcal{C}_i$. The canonical position of the camera in $\mathcal{C}_i$ is located along the $z$ axis.}
  \label{fig:intermediateFrame}
\end{figure}

As illustrated in Figure~\ref{fig:intermediateFrame}, we start by introducing an intermediary frame $\mathcal{C}_i$ for each camera which is centered in the body frame but oriented such that its $z$-axis aligns with the bar connecting the $i$-th camera in canonical position (i.e. without the influence of gravity). Let $(\mathbf{v}_i,\mathbf{V}_i)$ still denote the canonical extrinsics. The rotation matrix from $\mathcal{C}_i$ to $\mathcal{B}$ is given by
\vspace{-0.1 cm}
\begin{equation}
  \mathbf{B}_i = \left[\begin{matrix}
    \mathbf{b}_{1i} & \mathbf{b}_{2i} & \frac{\mathbf{v}_i}{\|\mathbf{v}_i\|}
  \end{matrix}\right],
\end{equation}
where $\mathbf{b}_{1i}$ and $\mathbf{b}_{2i}$ are the eigenvectors of $\mathbf{v}_i\mathbf{v}_i^T$ corresponding to the zero eigenvalues, and thus orthogonal to $\mathbf{v}_i$. The canonical extrinsic transformation of camera $i$ with respect to $\mathcal{C}_i$ is now given by
\vspace{-0.3 cm}
\begin{eqnarray}
\mathbf{q}_i & = & \left[\begin{matrix}0 & 0 & \|\mathbf{v}_i\|\end{matrix}\right]^T \ \\
\mathbf{Q}_i & = & \mathbf{B}_i^T\mathbf{V}_i. 
\end{eqnarray}

\vspace{-0.2 cm}
The final step consists of introducing the cantilever model to express the deformation of the bar. We start by transforming gravity into frame $\mathcal{C}_i$, i.e. $^{\mathcal{C}_i}\mathbf{g} = \mathbf{B}_i^T {^{\mathcal{B}}\mathbf{g}}$. This acceleration will now act on the point mass $m_i$ located at the tip of the bar and cause bar deformations. According to the cantilever model\cite{Hibbeler2017,Shan2015}, the third component and thus squeezing or stretching of the bar is ignored. The other two components cause the following deformations inside $\mathcal{C}_i$:
\begin{itemize}
    \item The first component of $^{\mathcal{C}_i}\mathbf{g}$ causes a displacement along $x$ and a rotation about $y$ given by
    \begin{eqnarray}
      \delta_{ix}({^{\mathcal{B}}}\mathbf{g}) & = & \frac{m_i \cdot \mathbf{b}_{1i}^T {^{\mathcal{B}}\mathbf{g}} \cdot \|\mathbf{v}_i\|^3}{3 \cdot E_iI_i} \\
      \theta_{iy}({^{\mathcal{B}}}\mathbf{g}) & = & \frac{m_i \cdot \mathbf{b}_{1i}^T {^{\mathcal{B}}\mathbf{g}} \cdot \|\mathbf{v}_i\|^2}{2 \cdot E_iI_i}.
    \end{eqnarray}
    \item The second component of $^{\mathcal{C}_i}\mathbf{g}$ causes a displacement along $y$ and a rotation about $x$ given by
    \begin{eqnarray}
      \delta_{iy}({^{\mathcal{B}}}\mathbf{g}) & = & \frac{m_i \cdot \mathbf{b}_{2i}^T {^{\mathcal{B}}\mathbf{g}} \cdot \|\mathbf{v}_i\|^3}{3 \cdot E_iI_i} \\
      \theta_{ix}({^{\mathcal{B}}}\mathbf{g}) & = & -\frac{m_i \cdot \mathbf{b}_{2i}^T {^{\mathcal{B}}}\mathbf{g} \cdot \|\mathbf{v}_i\|^2}{2 \cdot E_iI_i}.
    \end{eqnarray}
\end{itemize}

\begin{figure}[t]
  \centering
  \vspace{0.1 cm}
  \includegraphics[width=0.6\columnwidth]{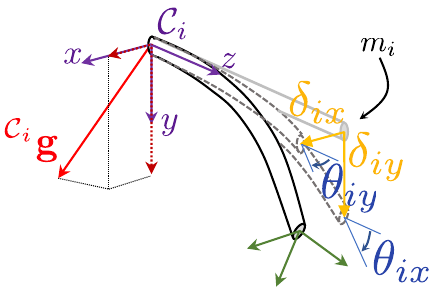}
  \vspace{-0.2 cm}
  \caption{Deformations $\delta_{ix}$, $\delta_{iy}$, $\theta_{ix}$ and $\theta_{iy}$ caused by the first and the second component of ${^{\mathcal{C}_i}}\mathbf{g}$.}
  \label{fig:deformations}
  \vspace{-0.5 cm}
\end{figure}
\vspace{-0.2 cm}
The geometry is depicted in Figure~\ref{fig:deformations}. Combined with the canonical transformations of the camera with respect to $\mathcal{C}_i$, and rotated back into the original body frame $\mathcal{B}$, we obtain the deformed extrinsic state as functions of gravity, given by
\vspace{-0.3 cm}
\begin{equation}
  \mathbf{w}_i = w_i(^{\mathcal{B}}\mathbf{g})  =  \mathbf{B}_i\
  \left[\begin{matrix}
    \delta_{ix}({^{\mathcal{B}}}\mathbf{g}) &   \delta_{iy}({^{\mathcal{B}}}\mathbf{g}) &   \|\mathbf{v}_i\|
  \end{matrix}\right]^T 
  \vspace{-0.4 cm}
\end{equation}
\begin{align}
      \mathbf{W}_i & = W_i(^{\mathcal{B}}\mathbf{g}) \nonumber \\
      & =  \mathbf{B}_i \mathbf{R}_x(\theta_{ix}({^{\mathcal{B}}}\mathbf{g})) \mathbf{R}_y(\theta_{iy}({^{\mathcal{B}}}\mathbf{g})) \mathbf{B}_i^T\mathbf{V}_i, 
  \label{eq:nonrigidextrinsics}
\end{align}
where $\mathbf{R}_x(\theta) = {\tiny\left[\begin{matrix}1 & 0 & 0 \\ 0 & \operatorname{cos}(\theta) & -\operatorname{sin}(\theta) \\ 0 & \operatorname{sin}(\theta) & \operatorname{cos}(\theta) \end{matrix}\right]}$, and $\mathbf{R}_y(\theta) = {\tiny\left[\begin{matrix}\operatorname{cos}(\theta) & 0 & \operatorname{sin}(\theta) \\ 0 & 1 & 0 \\ -\operatorname{sin}(\theta) & 0 & \operatorname{cos}(\theta) \end{matrix}\right]}$.
Note that these extrinsics are for the first body pose $\mathcal{B}$ given that we used the corresponding gravity vector. The extrinsics in the second pose are easily obtained by simply using the gravity in the second pose, i.e.
\vspace{-0.1 cm}
\begin{equation}
    \mathbf{w}'_i = w_i(^{\mathcal{B}'}\mathbf{g})\text{ and }\mathbf{W}'_i = W_i(^{\mathcal{B}'}\mathbf{g}).
\end{equation}

\subsection{Solving for both Pose and Gravity}
\label{sec:solvers}
\vspace{-0.1 cm}
A new generalized epipolar constraint for NR-MPCs in the form of \eqref{eq:generalizedepipolarconstraint} is obtained by using \eqref{eq:nonrigidextrinsics} inside the Pl\"ucker ray vectors, i.e.
\vspace{-0.2 cm}
\begin{eqnarray}
    \mathbf{l}_i & = & \left[\begin{matrix} W_{j_i}({^{\mathcal{B}}}\mathbf{g})\mathbf{f}_i \\ w_{j_i}({^{\mathcal{B}}}\mathbf{g}) \times \left(W_{j_i}({^{\mathcal{B}}}\mathbf{g})\mathbf{f}_i\right) \end{matrix} \right]  \\
    \mathbf{l}'_i & = & \left[\begin{matrix} W_{j_i}(\mathbf{R}^T{^{\mathcal{B}}}\mathbf{g})\mathbf{f}'_i \\ w_{j_i}(\mathbf{R}^T{^{\mathcal{B}}}\mathbf{g}) \times \left(W_{j_i}(\mathbf{R}^T{^{\mathcal{B}}}\mathbf{g})\mathbf{f}'_i\right) \end{matrix} \right]. 
\end{eqnarray}
Let us analyze the feasibility of a minimal solver~\cite{cox07,kukelova08solver}. The standard generalized relative pose problem has six unknowns, and thus requires at least six correspondences to be solved. The case of NR-MPCs is significantly harder owing to the addition of the latent gravity variable. Note that the norm of $\mathbf{g}$ is known, hence---in the minimal case---gravity is a non-linear function of only two unknowns. Also note that a single unknown gravity (e.g. ${^{\mathcal{B}}}\mathbf{g}$) is enough as the gravity in the second pose can be obtained by multiplication with $\mathbf{R}^T$. However, this does not make the problem easy. It still has eight degrees of freedom, both $w(\cdot)$ and $W(\cdot)$ are highly nonlinear functions of gravity, and the Pl\"ucker coordinates in the second pose additionally depend on the rotation. We therefore propose the following two solution strategies:
\begin{itemize}
\item We ignore the bendability of the bars and simply use the canonical extrinsic parameters $(\mathbf{v}_{j_i},\mathbf{V}_{j_i})$ in both poses in order to apply the six-point algorithm by Stewenius et al.~\cite{stewenius05} embedded into Ransac~\cite{fischler81}. An angular reprojection error threshold is used. The strategy succeeds if the bendability of the bar structure is not too high, and the inlier threshold angle is chosen sufficiently loose to account for the model inaccuracy. Given initial values for $\mathbf{R}$ and $\mathbf{t}$ and an initial inlier set, we then sample gravity vectors ${^{\mathcal{B}}}\mathbf{g}$ in the first pose and attempt robust non-linear refinement over the inlier set. The objective is indicated below. After the first round of nonlinear refinement is completed, we re-evaluate the number of inliers over the complete initial set of correspondences. If the number of inliers increases, we conclude with another round of non-linear refinement over the new inlier set. If the cardinality of the inlier set after the first round of optimization did not increase, we attempt again with another gravity sample. In practice, we iterate through six different gravity samples, which are pointing along the positive or negative $x$, $y$ or $z$ directions in $\mathcal{B}$.
\item We assume that an initial gravity measurement in the body frame can be obtained. Let those measurements be denoted ${^{\mathcal{B}}}\tilde{\mathbf{g}}$ and ${^{\mathcal{B}'}}\tilde{\mathbf{g}}$. This permits us to simply use the fixed Pl\"ucker coordinates
\begin{eqnarray}
    \mathbf{l}_i & = & \left[\begin{matrix} W_{j_i}({^{\mathcal{B}}}\tilde{\mathbf{g}})\mathbf{f}_i \\ w_{j_i}({^{\mathcal{B}}}\tilde{\mathbf{g}}) \times \left(W_{j_i}({^{\mathcal{B}}}\tilde{\mathbf{g}})\mathbf{f}_i\right) \end{matrix} \right]  \\
    \mathbf{l}'_i & = & \left[\begin{matrix} W_{j_i}({^{\mathcal{B}'}}\tilde{\mathbf{g}})\mathbf{f}'_i \\ w_{j_i}({^{\mathcal{B}'}}\tilde{\mathbf{g}}) \times \left(W_{j_i}({^{\mathcal{B}'}}\tilde{\mathbf{g}})\mathbf{f}'_i\right) \end{matrix} \right] 
\end{eqnarray}
to solve the problem. Note that the availability of a gravity measurement in each frame means that a directional correspondence between the two poses is given, and hence an initial guess for two degrees of freedom of the rotation is already given. Our strategy therefore consists of applying the four-point algorithm by Sweeney et al.~\cite{sweeney14} embedded into Ransac~\cite{fischler81}, followed by nonlinear optimization over the identified inlier set. This time, a tighter angular reprojection error threshold within ransac may be used. Note that it is also possible to work with gravity measurements inside one of the cameras, i.e. ${^{\mathcal{C}_i}}\tilde{\mathbf{g}}$. It can easily be converted into the body frame by optimizing ${^{\mathcal{B}}}\mathbf{g}$ such that---when combined with the resulting extrinsics $W_i({^{\mathcal{B}}}\mathbf{g})$---it agrees with the gravity measured in the camera frame. Using chordal distance, this optimization objective formally reads
\vspace{-0.1 cm}
\begin{equation}
\underset{ {^{\mathcal{B}}}\mathbf{g}  }{\operatorname{argmin}} \| W_i^T({^{\mathcal{B}}}\mathbf{g})\cdot{^{\mathcal{B}}}\mathbf{g} - {^{\mathcal{C}_i}}\tilde{\mathbf{g}}\|^2,\label{eq:gravityTrans}
\end{equation}
\vspace{-0.1 cm}
and we verified experimentally that it always converges if initialized from ${^{\mathcal{C}_i}}\tilde{\mathbf{g}}$.
\end{itemize}

The non-linear optimization objective is a two-view bundle adjustment problem over poses, points, and gravity. Let $\mathbf{p}_1,\ldots,\mathbf{p}_N$ be the 3D world points for our correspondences (expressed in the first body frame $\mathcal{B}$). Using angular errors, the objective is given by
\vspace{-0.3 cm}

\footnotesize
\begin{eqnarray}
\vspace{-0.3 cm}
      \underset{\mathbf{t},\mathbf{R},{^{\mathcal{B}}}\mathbf{g},\mathbf{p}_i}{\operatorname{argmin}} \sum_{i=1}^{N}
    \left\{
    h\left(\angle\left(
    \mathbf{f}_i,
    W_{j_i}^T({^{\mathcal{B}}}\mathbf{g})\left\{\mathbf{p}_i-w_{j_i}({^{\mathcal{B}}}\mathbf{g})\right\}
    \right)\right)\right. + \nonumber \\
   \left. h\left(\angle\left(
    \mathbf{f}'_i,
    W_{j_i}^T(\mathbf{R}^T{^{\mathcal{B}}}\mathbf{g})\left\{\mathbf{R}^T\left[\mathbf{p}_i-\mathbf{t}\right]-w_{j_i}(\mathbf{R}^T{^{\mathcal{B}}}\mathbf{g})\right\}
    \right)\right)
    \right\}, 
\end{eqnarray}
\normalsize

\noindent where $h(\cdot)$ represents the robust Huber-norm. Note that---for the sake of a simplified notation---the above objective integrates over all correspondences, while---as mentioned in the preceding algorithm outlines---in practice we will only consider subsets as identified via comparison against an inlier threshold. Note furthermore that ${^{\mathcal{B}}}\mathbf{g}$ remains parameterized as a minimal function of an azimuth and an elevation angle throughout all optimization objectives, and the rotation is parametrized minimally using Cayley parameters~\cite{cayley46}.
\begin{figure*}[t]
\vspace{-0.2 cm}
  \includegraphics[width=1.0\textwidth]{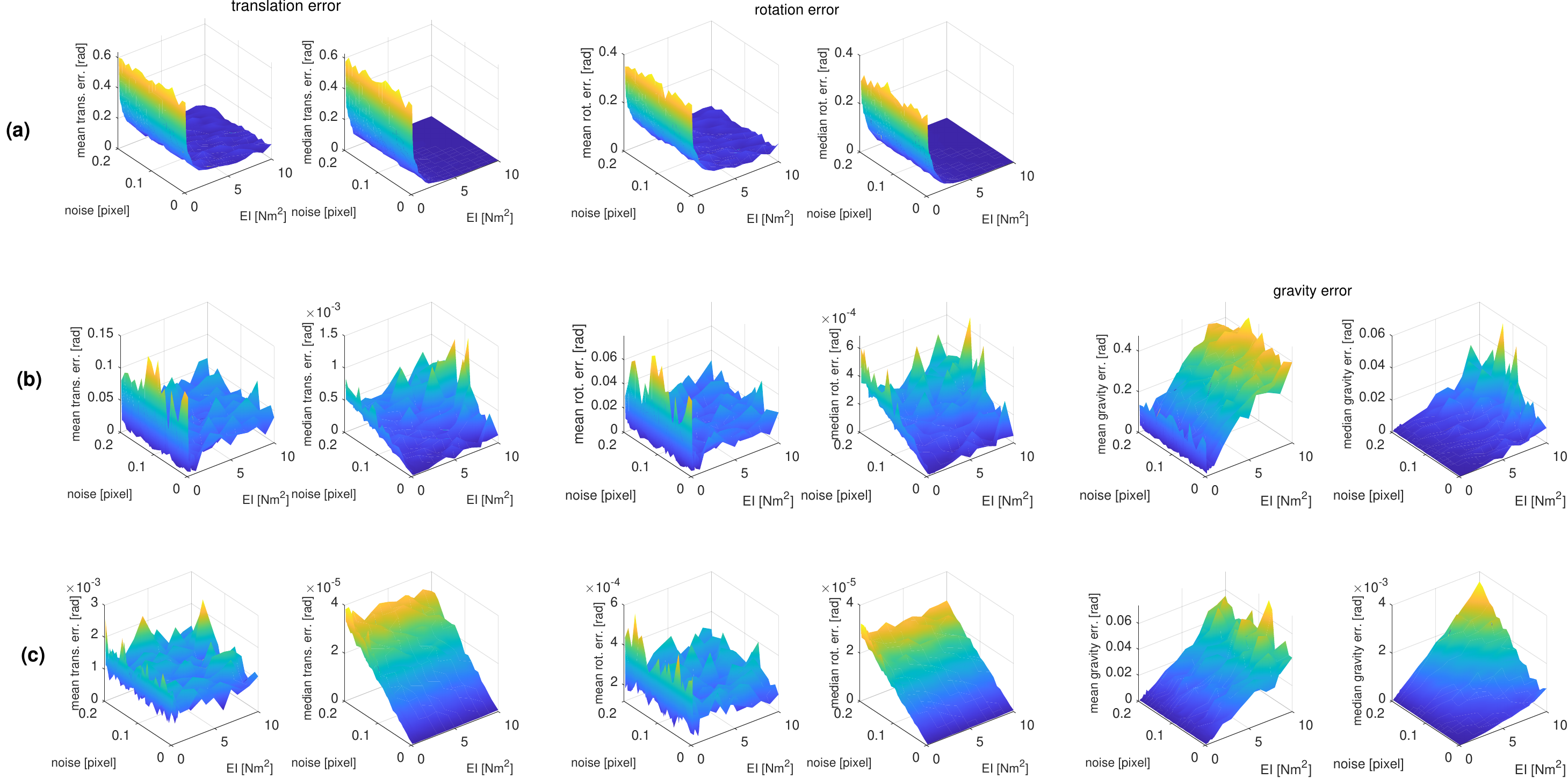}
\vspace{-0.4 cm}
  \caption{Translation, rotation and gravity errors for
  (a) 6pt-nonbending, (b) 6pt-bending, and (c) 4pt-bending. The plots show both mean and median errors over noise and bar rigidity.}
  \label{fig:4cams_allAlgorithms_errors}
  \vspace{-0.6 cm}
\end{figure*}

\noindent\textbf{Degenerate cases:} An in-depth discussion of degenerate cases would go beyond the scope of this work, but a few important ones can be immediately observed:
\begin{itemize}
    \item At least two cameras need to be used in order to make scale observable.
    \item If the system undergoes pure translation, scale remains unobservable irrespective of the number of cameras. This is similar to the rigid generalized relative pose problem.
    \item If the MPC has only two cameras and their bars are parallel, gravity will become unobservable if the system enters a pure translation along the direction that is orthogonal to both axes as well as gravity.
\end{itemize}

%% file: sec/4_syntheticExp.tex
\vspace{-0.2 cm}
\section{Synthetic Experiments}
\vspace{-0.2 cm}
We conduct rigorous synthetic experiments to compare the two solution strategies suggested in Section~\ref{sec:solvers}. As a further baseline implementation, we considered a third solution approach which simply ignores bending and assumes canonical extrinsic transformation throughout ransac and nonlinear optimization. Note that we do not expect this algorithm to be competitive, but it is included in order to underline the error committed by simply applying the current state-of-the-art technique in a case where non-rigidity occurs. For simplicity, we denote the above three approaches as \textit{6pt-bending}, \textit{4pt-bending}, and \textit{6pt-nonbending}, respectively.
All results are obtained by C++ implementations running on an Intel Core i5-8250U 8-core CPU. The virtual cameras are assumed to have VGA resolution and focal lengths of 800 pixels, and---in the case of 4-camera setups---are assumed to be placed at the tips of a regular cross-shape while the suspended body frame is assumed to lie at the center. Random 2D-2D correspondences are generated by distributing 3D points in a [-2,2]$\times$[-2,2]$\times$[4,8] volume in front of each camera in view 1, and then transforming those points into the second view-point.
The connections between the cameras and the body center are simulated by cantilever bars with a length of 0.4 m and a point mass of 0.1 kg at their tip, respectively. The nominal rigidity EI of the bar is set to be 0.1 Nm$^2$, a value for which the bar will have considerable nonrigidity. Moreover, EI is varied between 0.1 and 10 to see the impact of different bar rigidities. Different levels of Gaussian noise are added to the image and gravity measurements, and outliers are added by randomizing the direction vectors of a fraction of the correspondences. The number of correspondences for all experiments is 1000, and all errors are calculated based on 1000 runs.
Errors are calculated and compared for different noise levels, EI values, outlier fractions, as well as gravity measurement noise (for \textit{4pt-bending}). $\mathbf{t}_{gt}$, $\mathbf{t}$, $\mathbf{R}_{gt}$, $\mathbf{R}$, $\mathbf{g}_{gt}$, and $\mathbf{g}$ denote the ground truth and estimated translation, rotation and gravity. The rotation error is given by the norm of the Rodrigues representation of $\|\mathbf{R}_{gt}^T\mathbf{R}-\mathbf{I}\|$, which is the angle of the residual rotation expressed in radians. The translation error is given by $\|\mathbf{t}-\mathbf{t}_{gt}\|$, and the gravity error is expressed as the angle between $\mathbf{g}$ and $\mathbf{g}_{gt}$. Errors in translation are only evaluated according to their direction given that the scale of the translation is unobservable by MPCs in scenarios such as purely translational displacements. Convergence failures caused by degenerate cases or cases where the wrong model usage by \textit{6pt-nonbending} or \textit{6pt-bending} has detrimental impact on the quality of the result are removed by setting a threshold of 0.2 rad and 0.5 rad to translation and rotation errors, respectively.
\vspace{-0.1 cm}
\begin{figure}[t]
\centering
  \includegraphics[width=0.8\columnwidth]{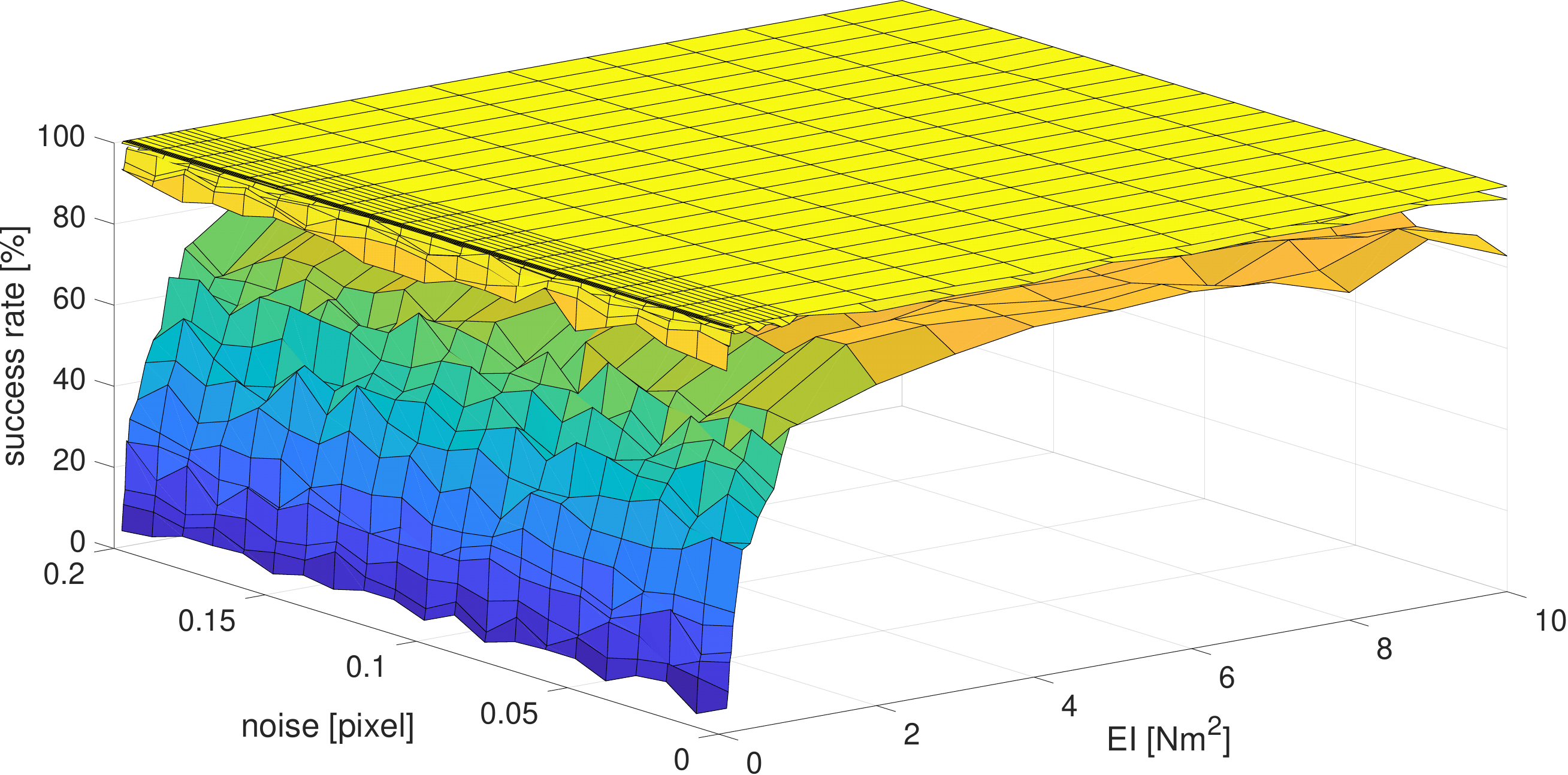}
  \vspace{-0.2 cm}
  \caption{Success rate for \textit{6pt-nonbending} (lowest layer), \textit{6pt-bending} (middle layer), and \textit{4pt-bending} (highest layer).}
  \label{fig:success_rate_4cams}
  \vspace{-0.6 cm}
\end{figure}
\subsection{Noise-EI Experiments}

\begin{figure}[b]
\centering
\vspace{-0.6 cm}
  \includegraphics[width=1.0\columnwidth]{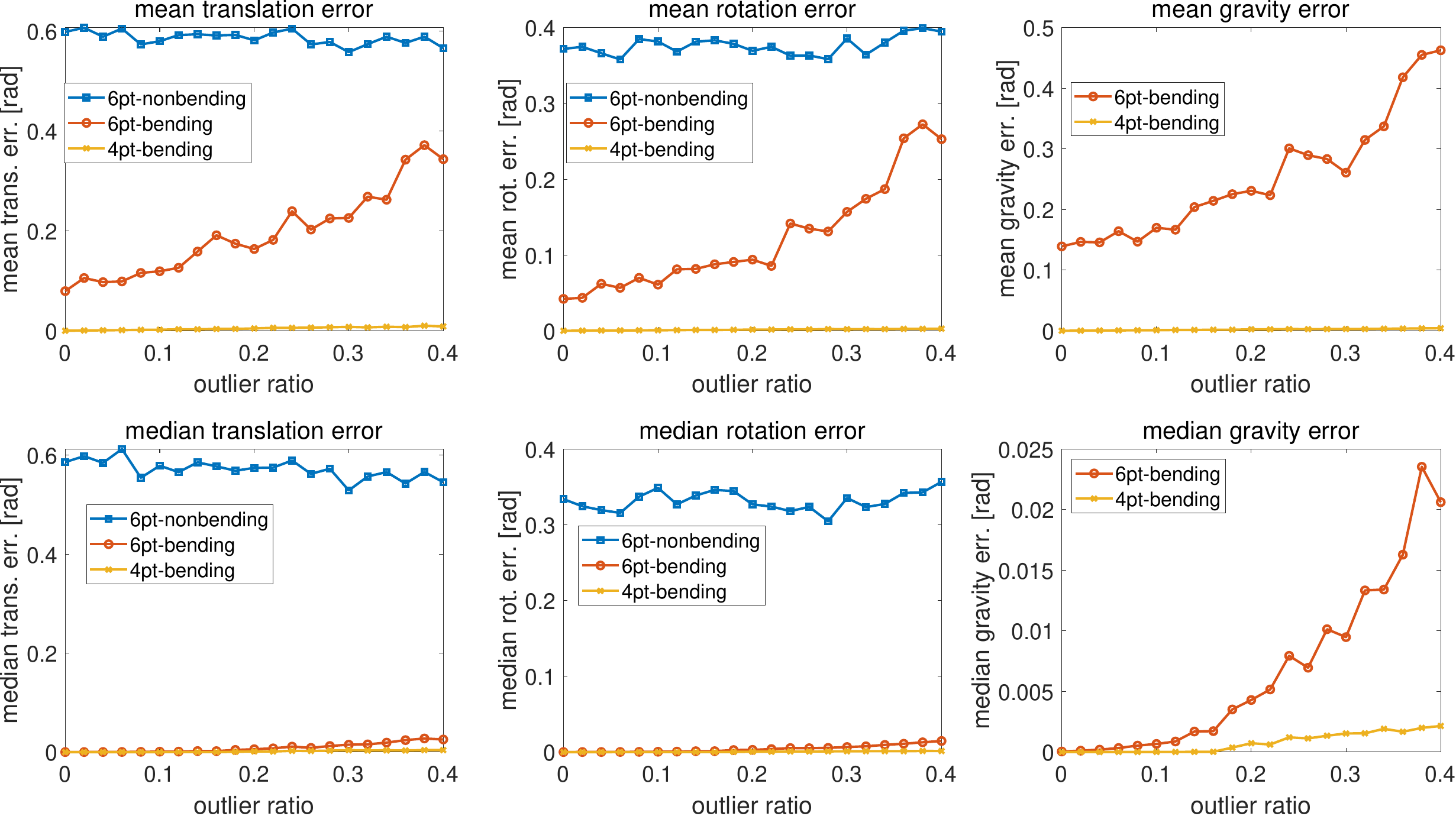}
\vspace{-0.6 cm}
  \caption{Resilience against increasing outlier ratios.}
  \label{fig:outlierTest_4cams}
  \vspace{-0.4 cm}
\end{figure}

Figure \ref{fig:4cams_allAlgorithms_errors} shows the mean and median translation, rotation and gravity errors for each approach as a function of pixel noise and bar rigidity. The noise is varied from 0.0 to 0.2 pixels, and EI is varied from 0.1 to 10. The noise in the gravity measurement is kept at 0.01, and the outlier ratio is set to 0.05.  It is worth mentioning here that there is no gravity error in \textit{6pt-nonbending} as this one simply does not optimize gravity. It is found that \textit{4pt-bending} has the least errors in translation and rotation among all three algorithms. \textit{4pt-bending} and \textit{6pt-bending} remain generally applicable even for increased rigidities as translation and rotation errors remain low and are mainly a function of pixel noise. However, the gravity error presents additional errors as the bar rigidity increases. Lastly, it is interesting to see that translation and rotation errors for \textit{6pt-nonbending} decrease as the rigidity increases, which points out the fact that the latter algorithm is in fact designed for the rigid case.

Figure \ref{fig:success_rate_4cams} additionally shows the success rate for all algorithms when subjected to the threshold for detecting gross errors. \textit{4pt-bending} has almost 100 percent success rate throughout the experiments. \textit{6pt-bending} has lower success rate than \textit{4pt-bending}, but still significantly higher success rate than \textit{6pt-nonbending}. The success rate of \textit{6pt-nonbending} is only several percent when EI is low, but increases greatly when EI increases and reaches equivalent success rate to \textit{6pt-bending} when the bar is quite rigid.

\subsection{Robustness to Outliers and Noise in Gravity}

\begin{figure}[t]
  \includegraphics[width=1.0\columnwidth]{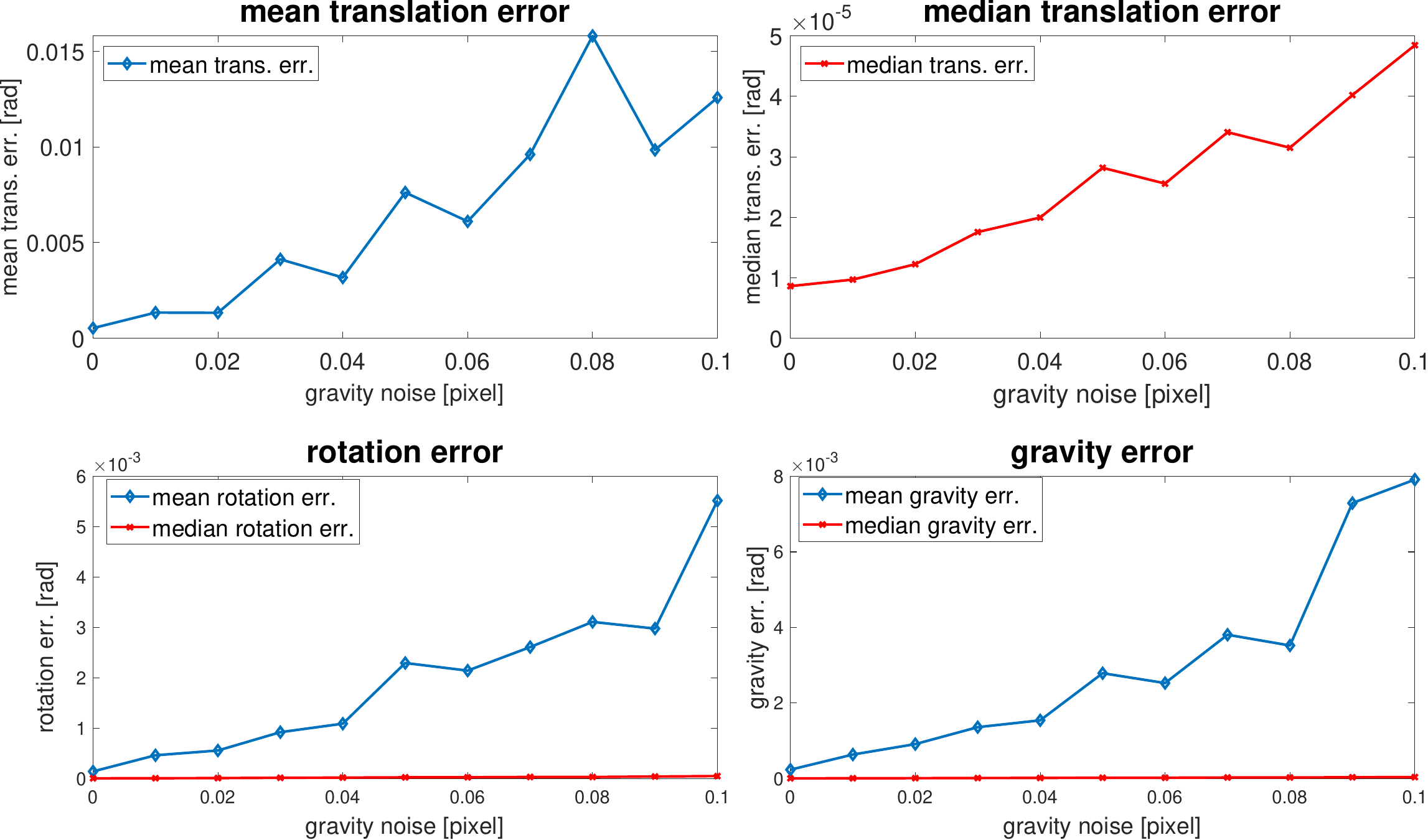}
\vspace{-0.5 cm}
  \caption{Resilience with respect to noise in initial gravity measurements.}
  \vspace{-1.2 cm}  
  \label{fig:noiseGravity_4cams}
\end{figure}

Figure \ref{fig:outlierTest_4cams} shows errors for varying outlier ratios between 0 and 0.4. 
As can be observed from Figure \ref{fig:outlierTest_4cams}, all errors gently increase along with larger outlier ratios, but \textit{4pt-bending} again presents the smallest errors, and \textit{6pt-nonbending} the largest.
Figure \ref{fig:noiseGravity_4cams} shows the influence of noise in the original gravity measurement for \textit{4pt-bending}. The gravity measurements are assumed to be direction vectors on the camera unit-sphere, and transformed for averaging into the body frame using \eqref{eq:gravityTrans}. Gravity noise is hence also expressed in pixels, and varied from 0 to 0.1 pixel. The outlier ratio is again fixed to 0.05, and EI is kept at 0.1 Nm$^2$. The mean and median errors of translation, rotation, and gravity increase gently with the noise in the observed gravity vectors.

\begin{figure*}[t]
  \includegraphics[width=1.0\textwidth]{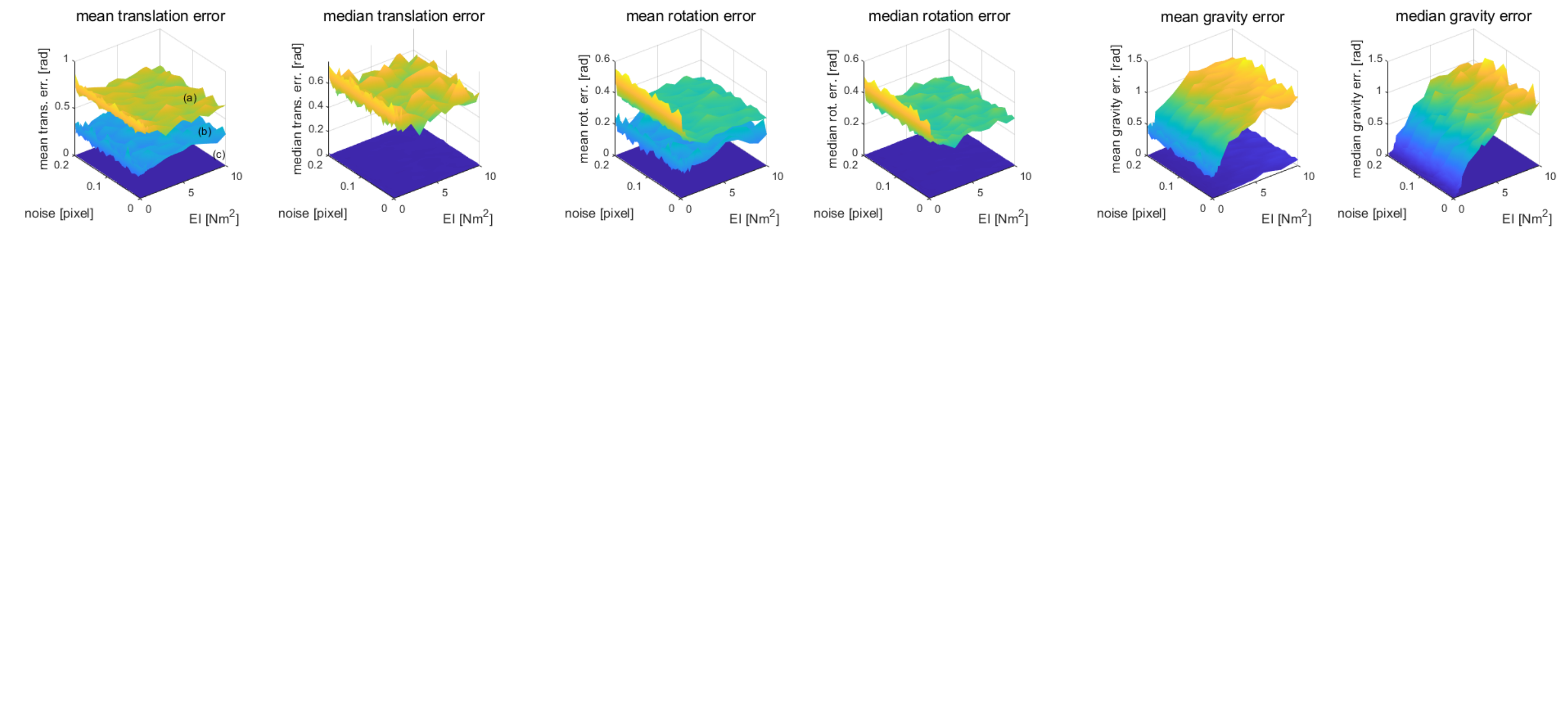}
  \vspace{-6 cm}
  \caption{Translation, rotation and gravity errors for
  (a) 6pt-nonbending, (b) 6pt-bending, and (c) 4pt-bending. The plots show both mean and median errors over noise and bar rigidity (2 cameras setup), with (c) lowest error values and (a) largest error values}
  \label{fig:2cams_allAlgorithms_errors}
  \vspace{-0.5 cm}
\end{figure*}
\vspace{-0.2 cm}
\begin{figure}[b]
\vspace{-0.8 cm}
  \includegraphics[width=1.02\columnwidth]{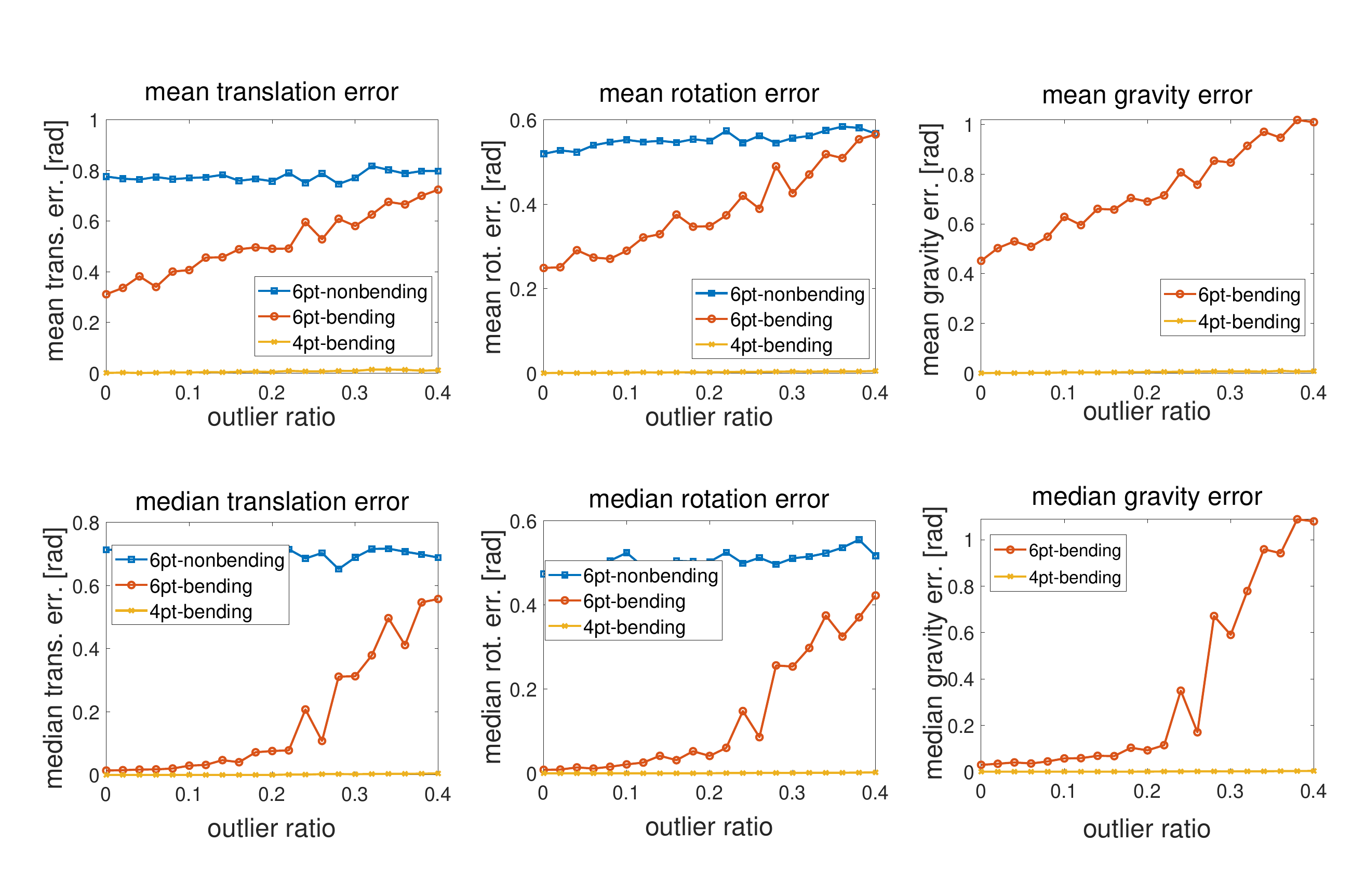}
\vspace{-0.9 cm}
  \caption{Resilience against outliers (2 camera setup).}
\vspace{-0.4 cm}
  \label{fig:outlierTest_2cams}
\end{figure}

\begin{figure}[b]
  \includegraphics[width=1.02\columnwidth]{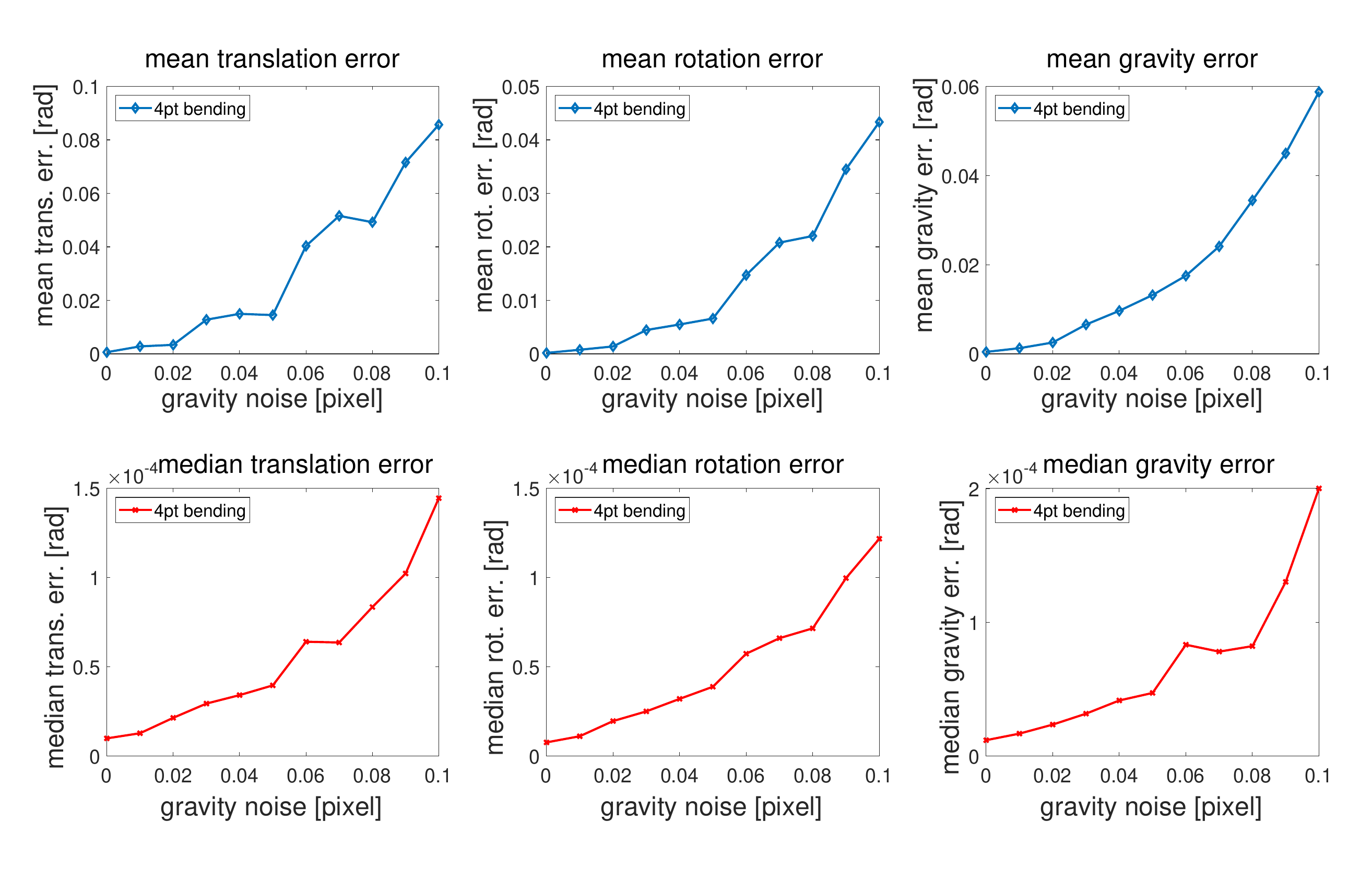}
\vspace{-0.7 cm}
  \caption{Resilience with respect to noise in initial gravity measurements (2 camera setup).}
  \label{fig:noiseGravity_2cams}
\end{figure}

\subsection{Efficiency and impact of camera number}
\vspace{-0.1 cm}
\textit{4pt-bending} has the best time efficiency among the three algorithms. The mean average running time over all experiments is around 0.05 s. The running time for \textit{6pt-nonbending} is very time-consuming as ransac has to exhaustively sample. A single run takes between 2 to 5 seconds to finish. The running time for \textit{6pt-bending} is lower, but normally still takes around 1 s to converge.  
As expected, the running time of \textit{6pt-nonbending} and \textit{6pt-bending} decreases quickly as the rigidity of the bar increases.

Further experiments are conducted on a 2-camera setup. Figure \ref{fig:2cams_allAlgorithms_errors}, Figure \ref{fig:outlierTest_2cams} and Figure \ref{fig:noiseGravity_2cams}
showed the respective results in terms of noise resilience, outlier resilience, and resilience with respect to noise in the gravity measurement. The results share a similar trend to the 4-camera system apart from a slight difference in the magnitude of the errors, thus confirming the correct functionality even in the case of only two cameras. In general, the number of cameras is arbitrary.
\vspace{-0.3 cm}

%% file: sec/5_realExp.tex
\vspace{-0.1 cm}
\section{Results on Real Data}
\vspace{-0.1 cm}
In order to demonstrate the validity of our findings, we also conduct frame-to-frame relative pose estimation experiments on real data. We set up our own non-overlapping 2-camera system mounted on a stainless steel cantilever bar, and capture sets of images in an indoor environment. The center of the cameras and the cantilever bar are arranged coaxially. The weight of the cameras and the camera holder therefore acts approximately as point masses onto the bar tips. Ground truth for the indoor sequence is delivered by a highly accurate external motion capture system. The intrinsics and extrinsics of the camera system are carefully calibrated. Our implementations are based on the open-source libraries OpenGV~\cite{opengv} and OpenCV~\cite{opencv_library}. Figure \ref{fig:real_experiment_setup} illustrates the setup of the real experiments. The obtained results verified the applicability of our theory, as shown in Table \ref{tab:readl_exp}. 

 \begin{table}
  \centering
  \caption{Average errors of the real-world experiments.}
  \vspace{-0.1 cm}
  \begin{tabular}{@{}c|ccc@{}}
    \hline
     \multirow{2}{*}{method}& translation & rotation & gravity \\
     & (rad) & (rad) & (rad) \\
    \hline
    \textit{6pt-nonbending} & 0.5397 & 0.2501 & - \\
    \textit{6pt-bending} & 0.3384 & 0.0729& 0.9594 \\
    \textit{4pt-bending} & \textbf{0.1004} & \textbf{0.0278} & \textbf{0.6721}\\
    \hline
  \end{tabular}
  \label{tab:readl_exp}
  \vspace{-0.3 cm}
\end{table}

%% file: sec/6_discussion.tex
\vspace{-0.2 cm}
\section{Discussion}
\vspace{-0.1 cm}
The present paper proves the interesting fact that the multi-view constraints of a non-overlapping MPC are sufficient to constrain a parametric deformation model. Perhaps surprisingly, this allows us to not only improve accuracy, but also consider the sensor support structure as a passive IMU and observe latent variables such as the direction of gravity without employing additional sensors, making assumptions about the environment, or employing intra-camera correspondences. While we believe this to be an important discovery, the present work still has two limitations. First, though there are already practically relevant use-cases that may be described by the cantilever model (e.g. sensor rigs, UAVs), the model remains simple and may not suffice to express the deformation of a body composed of multiple arbitrarily connected parts. Our future work consists of investigating novel primal-dual optimization objectives that involve the much more generally applicable Finite Element Model (FEM). Second, the present study only covers the static case. In the more general dynamic case, arbitrary temporally varying forces will act on the body thus causing dynamic deformations over time. Our downstream studies investigate the substitution of time-continuous models into the physical force-deformation constraints to enable simultaneous temporal ego-motion, attitude, and acceleration estimation.
\vspace{-0.3cm}
\section{Acknowledgements}
\vspace{-0.1 cm}
The authors would like to thank the fund support from Natural Science Foundation of China (62250610225) and the Shanghai Natural Science Foundation(22dz1201900, 22ZR1441300), and the 3D printing support from ShanghaiTech SIST MachineShop on real experiment setup.